\documentclass[a4paper,conference]{IEEEtran}
\ifCLASSINFOpdf
   \usepackage[pdftex]{graphicx}
\else
\fi
\usepackage{amsmath}
\usepackage{array}

\usepackage{stfloats}
\usepackage{url}

\usepackage{xspace}
\usepackage{booktabs}
\usepackage[pagebackref,breaklinks,colorlinks]{hyperref}

\usepackage{xcolor}

\usepackage[capitalize]{cleveref}
\crefname{section}{Sec.}{Secs.}
\Crefname{section}{Section}{Sections}
\Crefname{table}{Table}{Tables}
\crefname{table}{Tab.}{Tabs.}

\begin{document}
%
\title{AmsterTime: A Visual Place Recognition Benchmark Dataset for Severe Domain Shift}



%
\author{\IEEEauthorblockN{Burak Yildiz\IEEEauthorrefmark{1},
Seyran Khademi\IEEEauthorrefmark{1},
Ronald Maria Siebes\IEEEauthorrefmark{2} and
Jan van Gemert\IEEEauthorrefmark{1}}
\IEEEauthorblockA{\IEEEauthorrefmark{1}Delft University of Technology, The Netherlands\\
Emails: \{b.yildiz,s.khademi,j.c.vangemert\}@tudelft.nl}
\IEEEauthorblockA{\IEEEauthorrefmark{2}Vrije Universiteit Amsterdam, The Netherlands\\
Email: r.m.siebes@vu.nl}}


\maketitle

\newcommand{\amstertime}{AmsterTime\xspace}
\newcommand{\amsterdam}{GLDv2-Amsterdam\xspace}

\begin{abstract}
We introduce \amstertime: a challenging dataset to benchmark visual place recognition (VPR) in presence of a severe domain shift. \amstertime  offers a collection of 2,500 well-curated images matching the same scene from a street view matched to historical archival image data from Amsterdam city. The image pairs capture the same place with different cameras, viewpoints, and appearances.  Unlike existing benchmark datasets, \amstertime is directly crowdsourced in a GIS navigation platform (Mapillary). We evaluate various baselines, including non-learning, supervised and self-supervised methods, pre-trained on different relevant datasets, for both verification and retrieval tasks. Our result credits the best accuracy to the ResNet-101 model pre-trained on the \textit{Landmarks} dataset for both verification and retrieval tasks by 84\% and 24\%, respectively. Additionally, a subset of Amsterdam landmarks is collected for feature evaluation in a classification task. Classification labels are further used to extract the visual explanations using Grad-CAM for inspection of the learned similar visuals in a deep metric learning models.
\end{abstract}


%
\IEEEpeerreviewmaketitle

\section{Introduction}
\label{sec:introduction}

Visual place recognition (VPR) involves inferring a geographical location of a single image with  broad applications in robotics, consumer photography, social media, and archival repositories. The question of "where was this photo taken?" is answered for a query image, by retrieving the most similar match from a geo-tagged gallery of images. Thus, VPR is conveniently formulated as a content-based image retrieval problem, where the image representation is key. 

The ideal image representation for VPR  maps all the images capturing the same place {\textit{close}} to each other, regardless of various viewpoints, illuminations, appearances, and capturing sensors. At the same time, similar places are mapped {\textit{far enough}} from all other images, in an n-dimensional latent space. \cite{Serajeh2020}, where the quality of the representation mapping is measured against precision and recall over all queries, commonly captured in mean average precision (mAP) \cite{revaud2019apgem}. In this conduct, ranking perfection is achieved once all images of the same place are ranked higher than all others in the gallery. These defined criteria for the VPR task, lend themselves to a image-similarity learning problem.  

\begin{figure}
\begin{center}
\includegraphics[width=\linewidth]{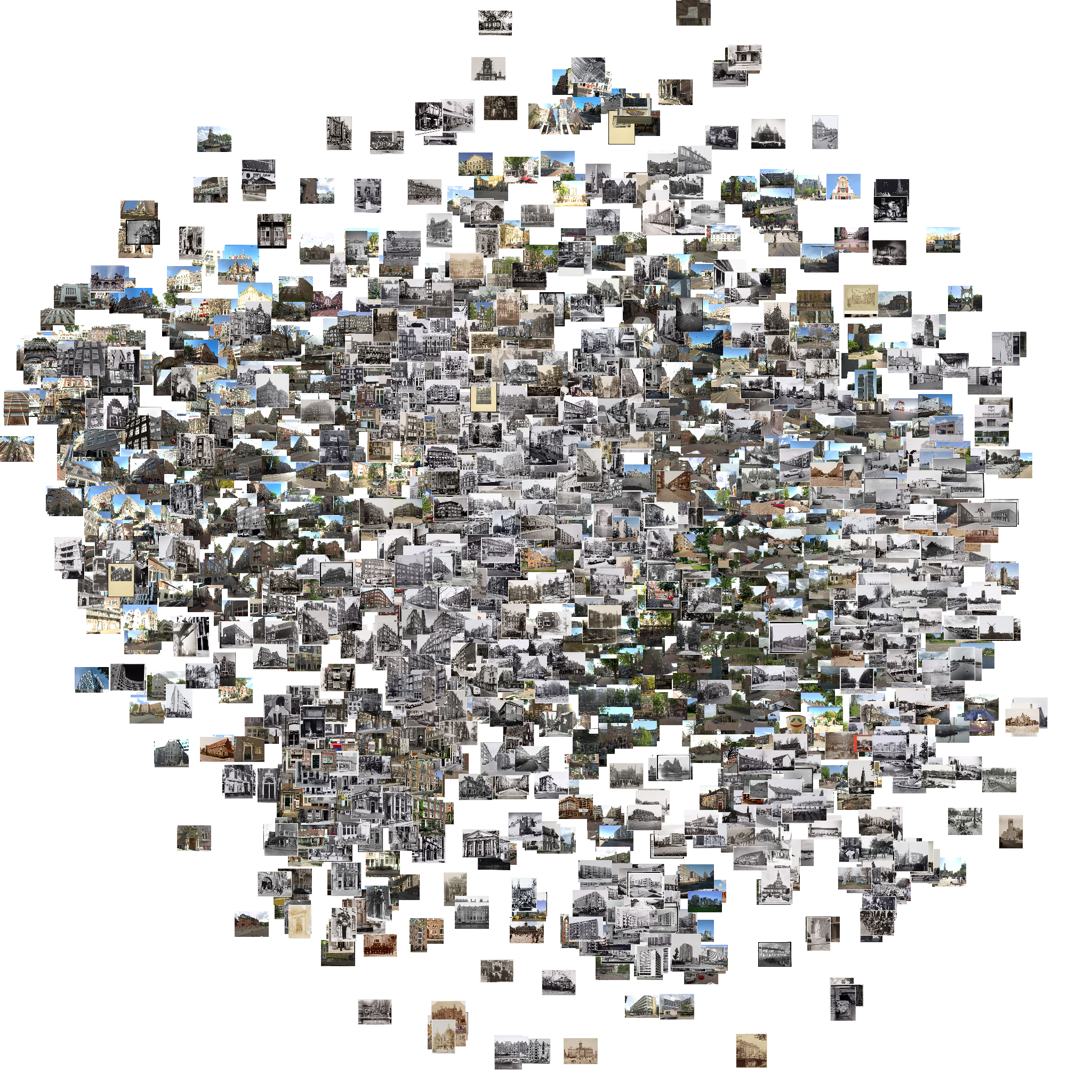}
\end{center}
\caption{The visualization of AmsterTime dataset using t-SNE}
\label{fig:tsne}
\end{figure}

In the past decade, deep similarity learning became a dominant framework by using (dis)similar image pairs to train convolutional neural networks (CNN)  such as Siamese \cite{Koch2015SiameseNN} and Triplet models \cite{Hoffer2015DeepML,balntas2016learning}.  At  inference time, the trained CNN model is used as a feature extractor for the image retrieval task. Similar to other learned image descriptors (features), the underlying relations in the training data determine similar visual elements. For instance, similar (positive) image pairs with different illuminations in the training, potentially leading to an illumination-agnostic model.
 
There exist numerous benchmark datasets for VPR, while none of them are directly crowd-sourced by retrieving similar visual places. A common approach to obtain positive images (most difficult) for VPR task is Geographic Information System based (GIS) annotations, e.g., from street view images with known GIS information \cite{arandjelovic2016netvlad}. GIS-based pairing, labels all the images with the same geographical altitude and latitude as similar, despite the fact that all the images taken from a single point do not share visual similarities, an example is orthogonal viewpoints. Others take images, taken by different people in social media or online photography platforms, from known landmarks such as Eiffel, Pyramids, etc \cite{gordo2016deep,weyand2020google}. A problem with the latter is the undesired bias towards popular geographical hotspots since many common architectural forms and typologies are excluded from the dataset. Other datasets use vehicle trajectories to capture the same scenes in different time frames ranging from seasonal to yearly intervals, tapering the scope of the VPR task to appearance-invariant learning and evaluation \cite{olid2018single,Chen2017}.  All of these semi-automated pair mining methods, even though efficient for learning relevant visual features, are either unfaithful to visual similarity notion or relatively facile to trustfully benchmark the VPR task.  In this paper, we introduce the first crowd-sourced benchmark dataset for the VPR task based on a visual search to match an archival query with street view images in the Mapillary navigation platform\footnote{\href{https://www.mapillary.com}{https://www.mapillary.com}}. In turn, all the matching pairs are verified by a human expert to verify the correct matches and evaluate the human competence in the VPR task for further references. The properties of our dataset referred to as \textit{AmsterTime}\footnote{This project is partly funded by \href{http://archimedial.eu/}{ArchiMediaL} project.} are summarized as:
 
\begin{itemize}
    \item 1200+ license-free images from the Amsterdam city Archive, representing urban places in the city of Amsterdam, captured in the past century by many photographers. 
    \item All archival queries are matched with street view images from Mapillary.
    \item All matches are verified by architectural historians and Amsterdam inhabitants.
    \item Image pairs are archival and street views capturing the same place with different cameras, time lags, structural changes, occlusion, viewpoint, appearance, and illuminations.
    \item The dataset exhibits a domain shift between query and the gallery due to significant difference between scanned archival and street view images.  
\end{itemize}

We embrace data scarcity as a realistic setting and we purposely limit \amstertime dataset for evaluating the VPR baselines rather than training. We also add visual similarity learning baselines with the latest self-supervision frameworks and visual inspection with Grad-CAM \cite{selvaraju2017grad} model to qualitatively evaluate the learned visual features. We list the contributions as:

\begin{enumerate}
     \item Various baselines including  recent self-supervised SimSiam~\cite{chen2020simsiam} model is evaluated on AmsterTime dataset.
     \item VGG-16~\cite{simonyan2014very} and ResNet-50~\cite{he2016deep} models are trained on a very large Google Landmarks  dataset~\cite{weyand2020google} for visual similarity learning with a self-supervised framework.
     \item Relevant landmarks from Amsterdam city are collected into a new classification dataset, from Google Landmarks dataset, to evaluate the learned similarity features, using class activation mapping frameworks such as Grad-CAM~\cite{selvaraju2017grad}.
     \item Visual explanations are generated using Grad-CAM model to inspect the visual similarities learned in the self-supervised models. 
\end{enumerate}

\begin{figure*}[t]
\centering
\includegraphics[width=1\linewidth]{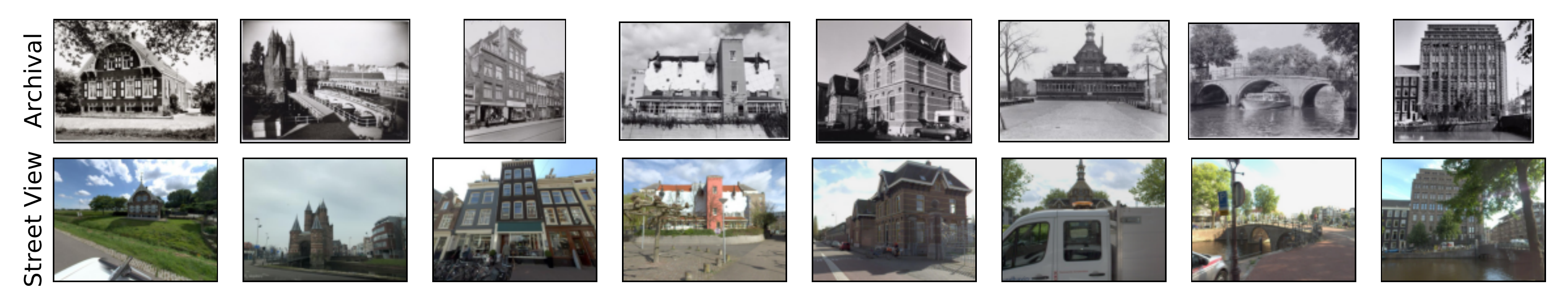}
\caption{Sample image pairs from \amstertime dataset. Challenges are extreme occlusions, view point changes, camera lens distortions, color changes.}
\label{fig:sample_images}
\end{figure*}

AmsterTime covers lifelong temporal coverage of Amsterdam city with severe domain shift between query and gallery which is uniquely challenging to benchmark VPR models as the baseline results indicate. t-SNE visualization of all the images in the dataset is given in \cref{fig:tsne} and some example image pairs are also given in \cref{fig:sample_images}. The dataset and the evaluation code are available at the project repository.\footnote{\href{https://github.com/seyrankhademi/AmsterTime}{https://github.com/seyrankhademi/AmsterTime}}
\section{Related Work}
\label{sec:related_work}

\begin{table}
\caption{Recent VPR datasets (left) with the corresponding data capturing and annotation medium (right). \amstertime combines two image domains to represent the same place.}
\label{tab:datasets}
\centering
\begin{tabular}{ll} \toprule
Datasets                                  & Imagery \\ \midrule
Oxford RobotCar~\cite{RobotCarDatasetIJRR}& Car Traverse \\
Berlin Kudamm~\cite{Chen2017}             & Train Traverse \\
Mapillary SLS~\cite{Warburg_2020_CVPR}    & Mapillary Street View \\
Pittsburgh-30k~\cite{Torii2015}           & Google Street View \\
Tokyo247~\cite{Torii2013}                 & Google Street View \\
Nordland~\cite{olid2018single}            & Train Traverse \\
Garden points~\cite{SunderhaufDSUM15}     & Car Traverse \\
\amstertime (ours)                        & Mapillary Street View + Archive \\ \bottomrule
\end{tabular}
\end{table}

\subsection{Datasets}
There are valuable survey papers that may be consulted for broader discussion over developments of VPR models and applications \cite{Masone2021,Zhang2020}. This work is based on \cite{Wang_2019_ICCV} that introduces unsupervised domain adaption and attention mechanism to solve the domain shift between query and gallery images in VPR task. Unlike \cite{Wang_2019_ICCV}, focusing on learning from large unpaired image sets from two domains of archival and street views, we develop a benchmark dataset of cross-domain image pairs to reliably evaluate learned image representations.

Among popular VPR datasets (\cref{tab:datasets}), the Berlin Kudamm dataset~\cite{Chen2017}, exhibit extreme viewpoint variation in the query and reference traverses. This dataset contains recurring and upfront dynamic objects which are uncommon to any other VPR dataset. Nordland dataset~\cite{olid2018single} sample images are one of the highly seasonally variant datasets and have manually introduced lateral viewpoint variation. Gardens Point dataset~\cite{SunderhaufDSUM15} images are presented here highly illumination variant and accompanied with lateral viewpoint variation. In contrast to these works, our dataset is unique in that it offers all the possible image variations including  viewpoints,  illuminations, appearances, and capturing sensors resulting in domain shift effect between the image pairs and thus extremely challenging.   

\subsection{Self-supervised representation learning}
In addition to the dataset, we investigate the performance of self-supervised similarity learning models for solving the VPR task, which is relevant because training data is scarce. 
In practice, constructing tuple training data and hard negative mining for deep visual similarity learning turned into a scalability bottleneck for suitable training datasets. Recently, Chen et al~\cite{chen2020simsiam} introduced a promising self-supervised framework based on Siamese networks that are trained only with positive image pairs, discarding altogether learning from dissimilar pairs.  We are inspired by \cite{chen2020simsiam}, that significantly reduces the combinatorial complexity of contrastive learning.

In general, self-supervised learning is used for task-agnostic representation learning \cite{caron2020unsupervised, he2020momentum, chen2020big, grill2020bootstrap, zbontar2021barlow, chen2020simsiam} commonly by a contrastive learning framework and Siamese networks. Interestingly, competitive performance is reported in the literature for self-supervised learning compare to the supervised learning models \cite{caron2020unsupervised, he2020momentum, chen2020big, grill2020bootstrap, zbontar2021barlow, chen2020simsiam}. Among the self-supervised models,  SimCLR~\cite{chen2020big} needs both negative and positive pairs with large batch sizes. 
While, SimSiam~\cite{chen2020simsiam} has an extra predictor module on one branch of its network which provides asymmetry and it prevents collapsing even with relatively small batch sizes in absence of negative pairs. Moreover, Barlow Twins~\cite{zbontar2021barlow} aims redundancy reduction in the representations by using a loss function on the cross-correlation matrix of the embedding which also prevents trivial solutions without the need for asymmetry in the Siamese setting.
\section{Crowdsourcing \amstertime }
\label{sec:dataset}

\begin{figure}[t]
\centering
\includegraphics[scale=0.5]{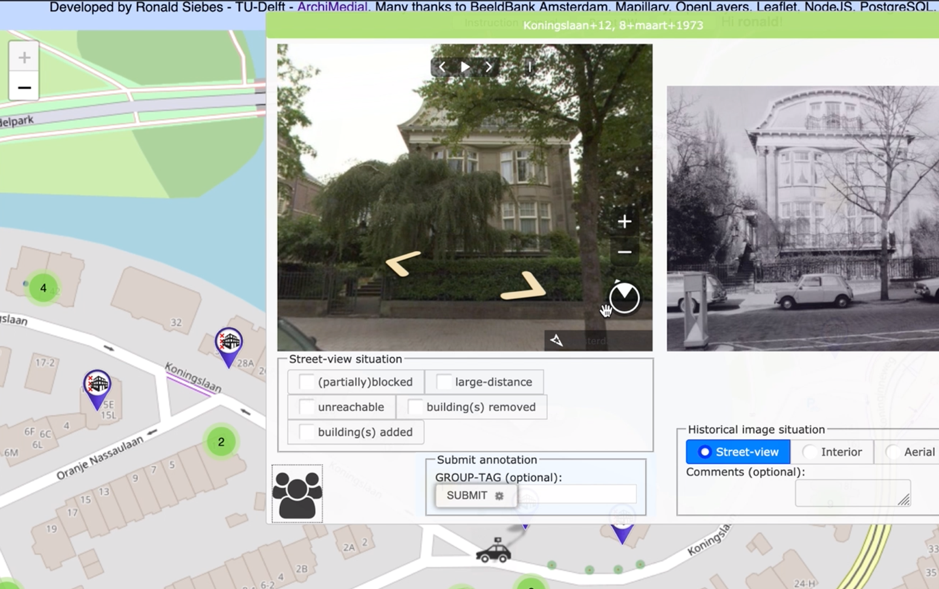}
\caption{Screenshot of the ArchiMediaL annotator app}
\label{fig:app}
\end{figure}

\begin{figure}
\begin{center}
\includegraphics[width=1\linewidth]{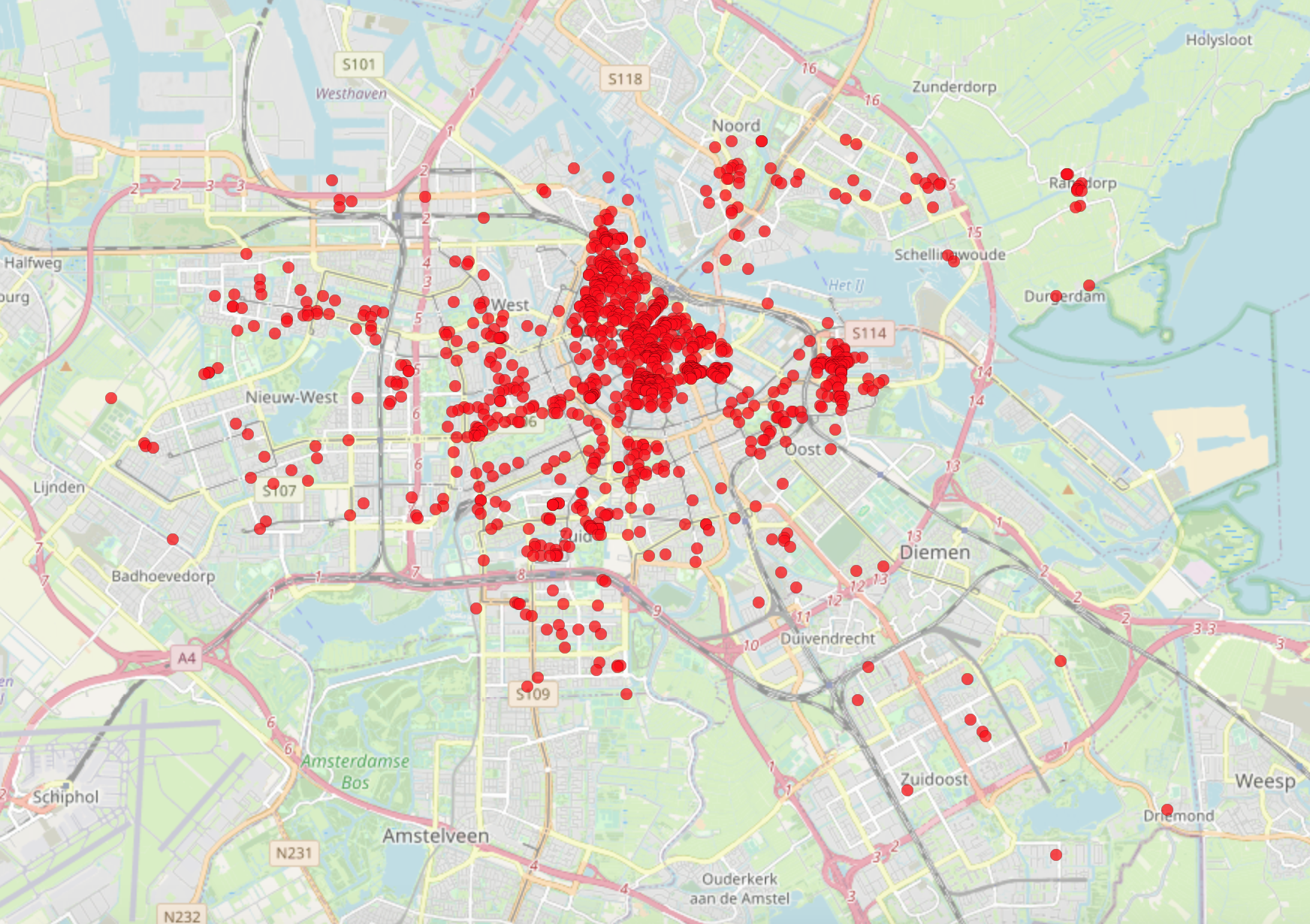}
\end{center}
\caption{The urban distribution of \amstertime dataset shows the concentration of data at the center of Amsterdam following the archival imagery pattern.}
\label{fig:AmsterMap}
\end{figure}

\subsection{Data Collection}

Crowdsourcing is a popular way to gather training and evaluation data for deep learning models. Well-known crowdsourcing platforms such as Amazon Mechanical Turk~\cite{crowston2012amazon} and AutoML \cite{AutoML}, are well suited for general crowdsourcing, yet, they are not adequate for our data collection application due to the complexity of the implementation of GIS layers. Therefore, we developed a custom crowdsourcing web application (\cref{fig:app}).
This annotation tool shows a participant the combination of an archival image and a 3D street-view navigator from the Mappilary platform. The navigator is positioned close to the expected location where the archival image is being taken, according to the available metadata, allowing the user to expand or zoom and match the archival and contemporary image in a game-like fashion. The tool also provides an evaluation interface, where administrators can manually verify or deny submissions. This task takes only a fraction of the time per image in comparison to the annotation task itself as it is a binary classification task of acceptance or rejection \cite{ArchiMediaL}. 

The selection of archival images originates from a fairly well-documented area of architectural and urban history (\cref{fig:AmsterMap}), in the Beeldbank repository of the Amsterdam City Archives\footnote{Beeldbank Stadsarchief Amsterdam. The Beeldbank contains several hundred thousand images taken in the streets of Amsterdam since the nineteenth century, among them many images of facades, buildings, and streets.   \url{https://archief.amsterdam/beeldbank/.}} – the world’s largest city archive. Moreover, the annotators are familiar with the place, from which the data is collected. 



\subsection{Benchmark Tasks}
\label{sub:tasks}
\amstertime  includes $1231$ matched archival and corresponding street view image pairs. We used these pairs to create both the {\textit{verification}} and {\textit{retrieval}} tasks that are closely related.

\textbf{Verification} is a binary classification (auxiliary) task to detect a pair of archival and street view images of the same place.  The verification task for \amstertime dataset has all of the crowdsourced image pairs as positive labeled, where the same number of negative samples are generated by randomly pairing archival and street view images summing up to a total of 2,462 pairs in the verification task.

\textbf{Retrieval} is the main task corresponding to VPR, in which a given query image is matched with a set of gallery images. For the retrieval task \amstertime dataset offers $1231$ query images where the leave-one-out set serves as the gallery images for each query.  
\section{Baseline Experiments and Results}
\label{sec:experiments}

\begin{table*}
\caption{Results for verification and retrieval tasks. The backbone architectures are given in the parentheses. The models are trained on the given dataset and evaluated on \amstertime dataset. SIFT and LIFT features are converted to 128 dimensional BoVW. The following three CNN architectures are ImageNet-pretrained models used only for evaluation. NetVLAD and AP-GeM architectures are also pre-trained models and used only for evaluation. Except the first SimSiam model, we trained the rest of them with self-supervision with the combination of 2 backbones and 3 datasets. The first SimSiam model is the pre-trained model from the original paper~\cite{chen2020simsiam} and used only for evaluation. Bold numbers denote the best scores for each column.}
\label{tab:results}
\centering
\begin{tabular}{llrrrrrrrr} \toprule
& & \multicolumn{5}{c}{\textbf{Verification}} & \multicolumn{3}{c}{\textbf{Retrieval}} \\ \cmidrule(r){3-7} \cmidrule(l){8-10}
\textbf{Method} & \textbf{Train Dataset} & \textbf{Precision} & \textbf{Recall} & \textbf{F1} & \textbf{Acc} & \textbf{ROC AUC} & \textbf{mAP} & \textbf{Top1} & \textbf{Top5} \\ \midrule
SIFT \cite{lowe2004distinctive} w/ BoVW & N/A & 0.57 & 0.65 & 0.61 & 0.58 & 0.61 & 0.03 & 0.01 & 0.04 \\
LIFT \cite{yi2016lift} w/ BoVW & Piccadilly & 0.56 & 0.60 & 0.58 & 0.57 & 0.59 & 0.03 & 0.01 & 0.04 \\ \midrule
VGG-16 \cite{simonyan2014very} & ImageNet & 0.75 & 0.63 & 0.68 & 0.71 & 0.78 & 0.18 & 0.13 & 0.23 \\
ResNet-50 \cite{he2016deep} & ImageNet & 0.63 & 0.66 & 0.65 & 0.64 & 0.69 & 0.06 & 0.04 & 0.08 \\
ResNet-101 \cite{he2016deep} & ImageNet & 0.63 & 0.67 & 0.65 & 0.64 & 0.69 & 0.05 & 0.03 & 0.07 \\ \midrule
NetVLAD (VGG-16) \cite{arandjelovic2016netvlad} & Pittsburgh250k & 0.83 & \textbf{0.80} & 0.82 & 0.82 & 0.90 & 0.26 & 0.17 & 0.33 \\
AP-GeM (ResNet-101) \cite{revaud2019apgem} & Landmarks & \textbf{0.88} & 0.78 & \textbf{0.83} & \textbf{0.84} & \textbf{0.92} & \textbf{0.35} & \textbf{0.24} & \textbf{0.48} \\ \midrule
SimSiam (ResNet-50) \cite{chen2020simsiam} & ImageNet & 0.75 & 0.76 & 0.75 & 0.75 & 0.83 & 0.19 & 0.12 & 0.26 \\
SimSiam (ResNet-50) & GLDv2 & 0.80 & 0.79 & 0.80 & 0.80 & 0.86 & 0.23 & 0.15 & 0.32 \\
SimSiam (ResNet-50) & \amstertime & 0.72 & 0.75 & 0.73 & 0.73 & 0.81 & 0.19 & 0.12 & 0.26 \\
SimSiam (VGG-16) & ImageNet & 0.63 & 0.72 & 0.67 & 0.65 & 0.71 & 0.10 & 0.06 & 0.14 \\
SimSiam (VGG-16) & GLDv2 & 0.63 & 0.77 & 0.70 & 0.66 & 0.75 & 0.12 & 0.07 & 0.18 \\
SimSiam (VGG-16) & \amstertime & 0.77 & 0.70 & 0.73 & 0.74 & 0.81 & 0.16 & 0.10 & 0.22 \\ \bottomrule
\end{tabular}
\end{table*}

\subsection{Experimental Setup}
\label{sub:setup}

We take the pairwise distance between two high-dimensional feature vectors corresponding to all the images in \amstertime. The average of all distance values generated by pairwise comparisons in the dataset is used as a threshold distance to classify positive (similar) or negative (dissimilar) pairs. None of the baseline models that we trained uses the dataset labels. Self-supervised models only use \amstertime images but does not use the pairing annotations.



We calculate mean average precision $mAP$, $Top1$ and $Top5$ accuracy metrics for retrieval task using $cosine$ distance. For a given query archival image, we first sort all street view images by the distance between the archival image and the street view images in ascending order then the metrics are calculated.

\subsection{Local Image Features}
\label{sub:local}
We investigated how local image features perform on \amstertime dataset. The SIFT~\cite{lowe2004distinctive} descriptors are used to extract local features and they were then aggregated into one global feature per image using bag-of-visual-words encoding (BoVW)~\cite{sivic2003bovw}. The process repeated for the descriptors extracted with the LIFT~\cite{yi2016lift} trained on \textit{Piccadilly} dataset~\cite{wilson2014piccadilly}. The bag size is chosen 128 which performs best among others. The results for the verification and retrieval tasks using BoVW are given in \cref{tab:results}. Accuracy for verification task for SIFT is $58\%$ ($8\%$ above the random baseline) and for LIFT $57\%$. 

\subsection{Off-the-shelf (pre-trained) CNN models}
\label{sub:cnn}
We evaluated the performance of image features extracted from commonly used CNN models pre-trained on different datasets and tasks including image classification and visual place recognition (VPR), on \amstertime dataset.

The models VGG-16~\cite{simonyan2014very}, ResNet-50~\cite{he2016deep}, and ResNet-101~\cite{arandjelovic2016netvlad} are pre-trained on ImageNet~\cite{deng2009imagenet} for image classification and used directly from PyTorch's library. The CNN models are only used for extracting features of the images in \amstertime dataset. The features are obtained from the last convolutional layer followed by a ReLU and a max-pooling layers for VGG-16 model and from adaptive average pooling layer for ResNet models. The features are then utilized to calculate scores for verification and retrieval tasks. The results are given in \cref{tab:results}. One noticeable point is that VGG-16 works better than both ResNet-50 and ResNet-101 on verification task. That margin is much bigger on retrieval task as VGG-16 has $13\%$ top-1 accuracy while ResNet-50 has $4\%$ and ResNet-101 has $3\%$ top-1 accuracy.

In the next step, we used  NetVLAD~\cite{arandjelovic2016netvlad} pre-trained on \textit{Pittsburgh250k}~\cite{Torii2013} for VPR task as a close match to our task of image retrieval. In addition, we evaluated  AP-GeM~\cite{revaud2019apgem}  pre-trained on \textit{Landmaks-clean} dataset~\cite{gordo2016deep} on \amstertime dataset. The NetVLAD model has a VGG-16 backbone while AP-GeM has ResNet-101 backbone. Neither of them are trained further than the publicly available model weights. As usual, the models are used to extract image features for both verification and retrieval task. AP-GeM and NetVLAD models result in $84\%$ and  $82\%$ accuracies on verification task, respectively. The pre-trained AP-GeM achieves the best performance among all the baselines as it leverages the largest training dataset which is very similar to images in \amstertime dataset.

\subsection{Self-supervised Baseline}
\label{sub:self}
Due to the limited size of \amstertime, self-supervision is a suitable option to exploit data without labels. SimSiam~\cite{chen2020simsiam} is a recent method that combines self-supervision and similarity learning without needing for neither negative samples nor large batches. We evaluated six SimSiam models with different data and architectures presented in the last row section in \cref{tab:results}. The list starts from ResNet-50 model trained on ImageNet\footnote{This model is used from SimSiam authors' shared models.}. We trained two more ResNet-50 models on Google Landmarks (GLDv2) and \amstertime datasets with same settings except that the batch size is $128$ in our trainings. Since ImageNet and GLDv2 are relatively large datasets and \amstertime is limited dataset, the model trained on GLDv2 is trained for $100$ epochs and the model trained on \amstertime dataset is trained for $10000$ epochs to equalize the number of used mini-batches during training. Moreover, three VGG-16\footnote{VGG-16 model architecture is with batch normalization which is used directly from PyTorch's library.} models are trained on the same datasets (ImageNet, GLDv2 and \amstertime) with the same settings as bare supervised VGG-16 has better results than ResNet-50 remarked in \cref{sub:cnn}. The models started the self-supervised training from scratch (with random parameters) and after the self-supervised training are completed, the trained models used as usual to extract features from the images in \amstertime dataset. The results are presented in \cref{tab:results}. Contrary to the better results for pre-trained VGG-16 model mentioned in \cref{sub:cnn}, ResNet-50 outperformed in self-supervised learning.

\subsection{Supervised Baseline}
\label{sub:supervision}

\begin{table}
\caption{Results of supervised trained ResNet-18 with triplet loss \cite{balntas2016learning} on \amstertime dataset. The numbers are merely to quantify the supervision gap compared to unsupervised models in Tab.~ \ref{tab:results}  }
\label{tab:supervised_results}
\centering
\begin{tabular}{rrrrrrr} \toprule
\multicolumn{4}{c}{\textbf{Verification}} & \multicolumn{3}{c}{\textbf{Retrieval}} \\
\cmidrule(r){1-4} \cmidrule(l){5-7}
\textbf{Precision} & \textbf{Recall} & \textbf{Acc} & \textbf{ROC AUC} & \textbf{mAP} & \textbf{Top1} & \textbf{Top5} \\ \midrule
0.85 & 0.89 & 0.87 & 0.93 & 0.42 & 0.30 & 0.53 \\ \bottomrule
\end{tabular}
\end{table}

To have a real supervised baseline and measure the supervision gap, we also trained a ResNet-18 model in Triplet setting \cite{balntas2016learning} with grand-truth pair labels. The dataset first divided into train and test with the ratio of $4:1$. Besides ground-truth (positive) pairs, equal number of pseudo negative pairs are randomly generated by pairing the archival and street-view images. The model trained for $90$ epochs with SGD optimizer with $128$ of batch size, $0.001$ learning rate (decayed by $0.1$ after each $30$ epochs), $0.9$ momentum and $0.00001$ weight decay. The model is then used to extract feature on the test set. The results for verification and retrieval are given in \cref{tab:supervised_results}. Due to the limited size of \amstertime dataset, the supervision gap is only around $7\%$ in {\textit{mAP}}. 



\section{Visual Explanations}
\label{sec:visualization}
We investigate the learned representation of the SimSiam~\cite{chen2020simsiam} on \amstertime dataset using   Grad-CAM~\cite{selvaraju2017grad}. 
Grad-CAM requires a classification layer at the end of CNN architecture while SimSiam-trained models trained on similarity learning. To adapt Grad-CAM, (1) we add a randomly initialized liner classifier at the end of the SimSiam-trained models, (2) train the newly added classifier on a curated similar dataset with class labels (landmarks). The parameters of SimSiam-trained models are frozen after training on \amstertime dataset. 

\subsection{Dataset for visualization}

\begin{figure}[t]
\centering
\includegraphics[width=1\linewidth]{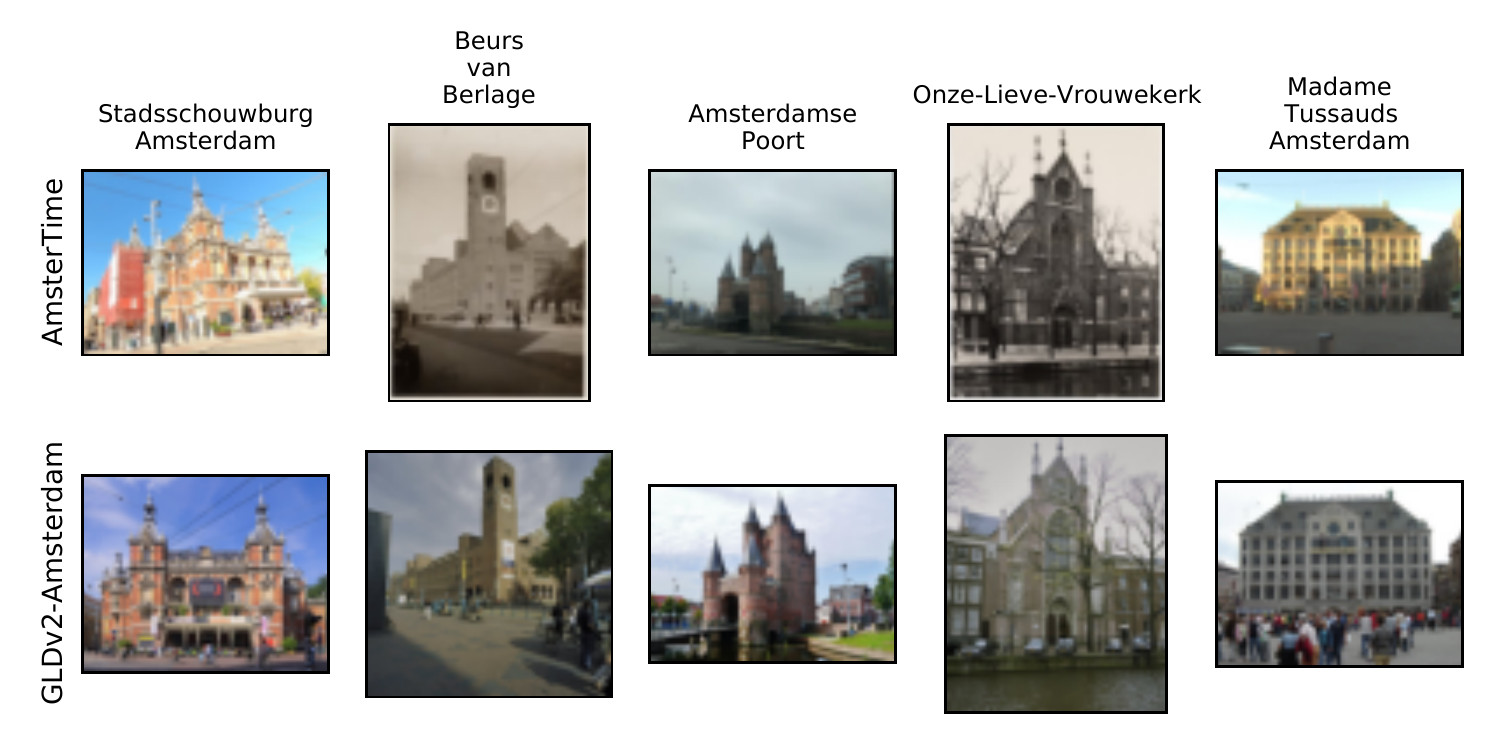}
\caption{Example of similar images for the same landmark classes in both \amstertime and \amsterdam datasets enables visualizing learned features with Grad-CAM.}
\label{fig:similar_images}
\end{figure}

\begin{figure}[t]
\centering
\includegraphics[width=1\linewidth]{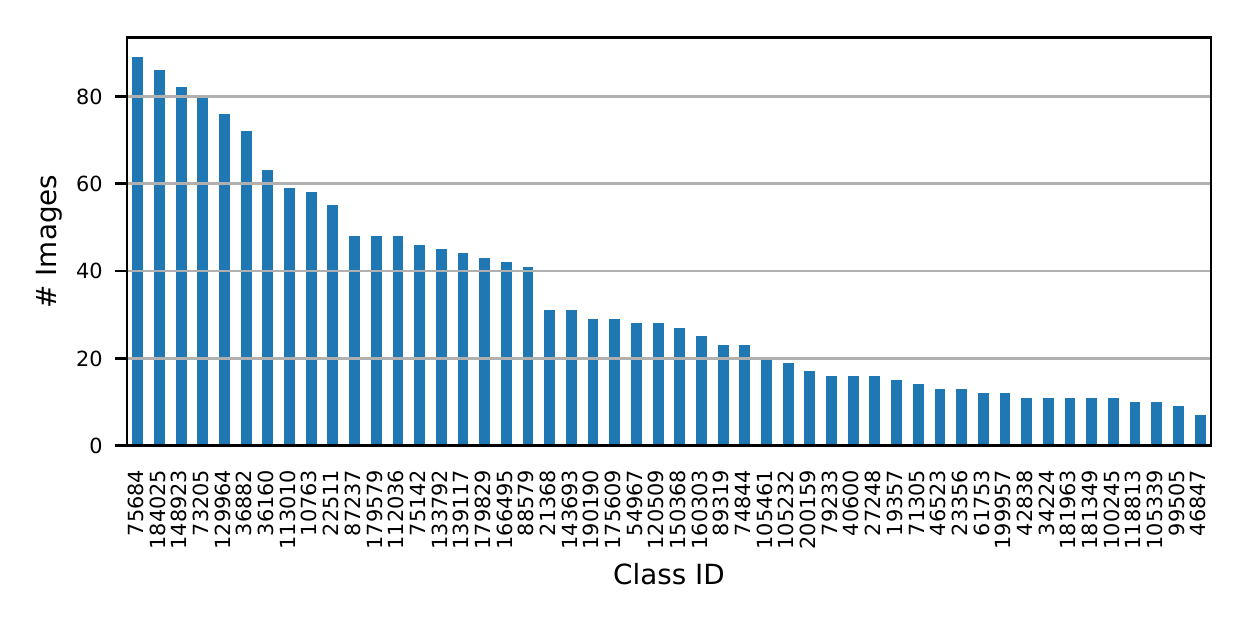}
\caption{Histogram of the number of images per selected 50 landmarks located in Amsterdam from Google Landmarks Dataset v2 shows that class distribution is unbalanced and solved simply by duplicating images in underrepresented classes.}
\label{fig:class_distribution}
\end{figure}

To facilitate visualization  we curated a subset of Google Landmarks dataset v2 (GLDv2)~\cite{weyand2020google} because it is semantically close to \amstertime. Particularly, $50$ landmarks are selected in GLDv2 which are located in the city of Amsterdam. Some of the hand-picked similar images have been given in \cref{fig:similar_images}. We will refer to this subset for the classification dataset as \textit{\amsterdam} hereafter. The histogram of class distribution of \amsterdam,  presented in  \cref{fig:class_distribution}, shows a highly skewed and imbalanced distribution. To alleviate the training suffering from the imbalanced dataset, of the linear classifier, underrepresented classes were simply duplicated \amsterdam.

\subsection{Training the Linear Classifier}
The linear classifier is trained based on \cite{chen2020simsiam}. A randomly-initialized linear classifier added to a frozen model is trained for $90$ {\textit{epochs}} with {\textit{batch size}}$=256$ using SGD  with the parameters of cosine-decay-scheduled, initial \textit{lr}$=30.0$ , \textit{weight decay}$=0$, and \textit{momentum}$=0.9$.

\subsection{Grad-CAM visualization}

\begin{figure}[t]
\centering
\includegraphics[width=1\linewidth]{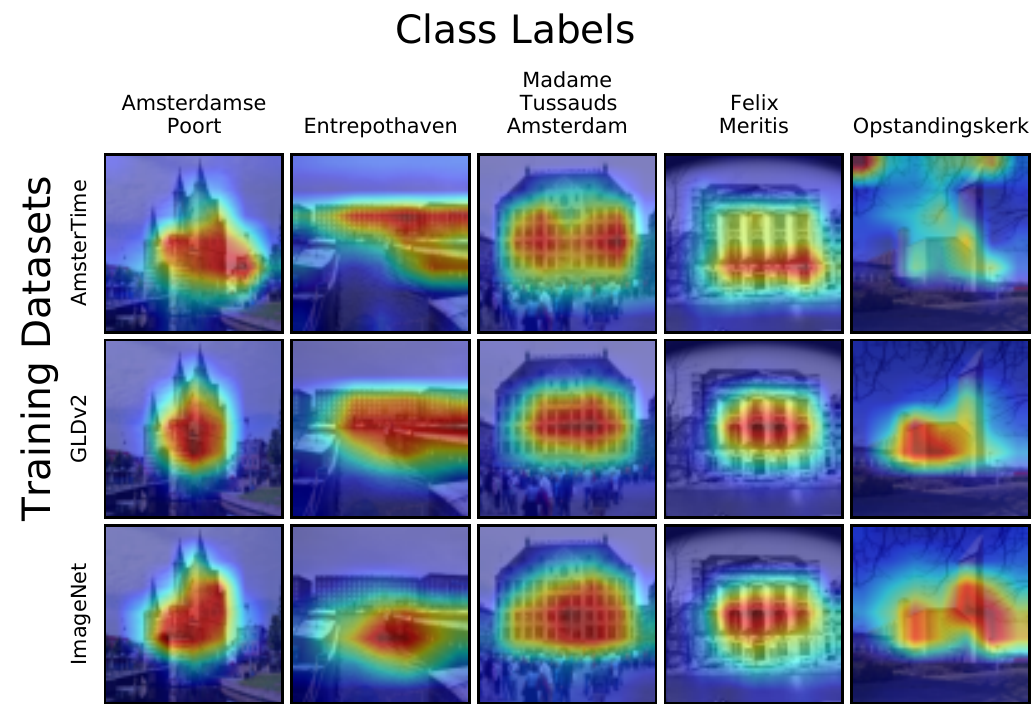}
\caption{Grad-CAM visualizations of three ResNet-50 models pre-trained with SimSiam~\cite{chen2020simsiam} on three different datasets given in the labels on the left. Top labels denotes both class activation for Grad-CAM and grand-truth class labels. The visualizations suggest that models learn the structure in the images.}
\label{fig:gradcam_simsiam}
\end{figure}

\begin{figure}[t]
\centering
\includegraphics[width=1\linewidth]{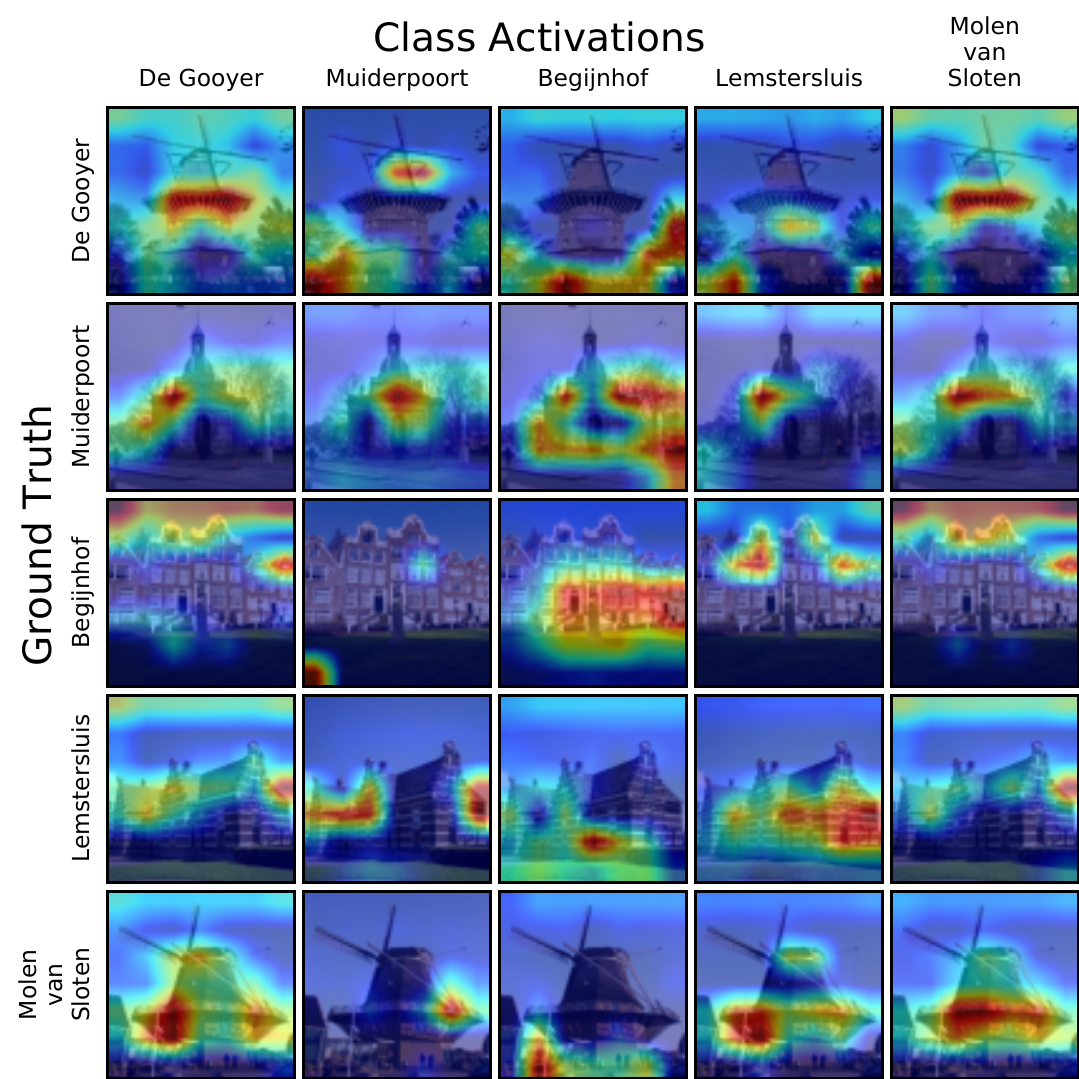}
\caption{Grad-CAM visualizations of ResNet-50 model pre-trained with SimSiam~\cite{chen2020simsiam} on \amstertime dataset. The visualizations on the diagonal and the intersection of De Gooyer and Molen van Sloten show that the model activates more when the activation class images are similar to the input images which indicates the model learned the structures in the images.}
\label{fig:gradcam_amstertime}
\end{figure}

We created two visualizations: The first is to show how the same models trained on different datasets learn different visual features, and the second one is to see how a model reacts to different class activations of Grad-CAM.

Firstly, we selected a subset among the intersection of correctly classified images in \amsterdam dataset by three ResNet-50 models pre-trained on ImageNet, GLDv2, and \amstertime with SimSiam self-supervision. The output of the last convolutional layer of the ResNet-50 models is visualized. The Grad-CAM visualizations for each of the selected images and for each of the models are created and imposed on the original images. In this setting, the grand-truth classes of the images are used as class activation to show the reactions of the models to the images w.r.t. grand-truth class activation. The visualizations are given in \cref{fig:gradcam_simsiam}. The models give more weight to the landmark objects in the images, indicating that the models correctly focus on landmark features.

To illustrate the reactions of the models in comparison with other class activations besides the grand-truths, we also created a Grad-CAM visualization matrix given in \cref{fig:gradcam_amstertime}. For this visualization matrix, we used the ResNet-50 model pre-trained on \amstertime dataset with SimSiam self-supervision. The visualizations for images w.r.t. grand-truth class activation appears on the diagonal. This matrix shows that the model relies on the landmark object once the class activation is either the grad-truth class or the class of the images with similar landmarks such as \textit{De Gooyer} and \textit{Molen van Sloten}.
\section{Conclusions}
We introduced \amstertime a reliable and challenging evaluation dataset with verification and retrieval benchmark tasks for visual place recognition. \amstertime dataset consists of $\sim$ 2500 archival and street view images matched by human annotators. Various image representation baselines including the local features, supervised and self-supervised models are tested on \amstertime. The results suggest that supervised model trained on a large and similar dataset of {\textit{Landmarks}} outperforms the self-supervised models.  Oblation studies are carried out using visual explanations to investigate the learned features confirming the quality of \amstertime dataset in learning relevant features despite its small size using self-supervised models. The code for this paper including the image features is available on a GitHub repository. 

\section{Acknowledgement}
This work is partially supported by Volkswagen Foundation under \href{http://archimedial.eu/}{ArchiMediaL} project. We show our gratitude to all the people who helped us to annotate data including Tino Mager, Beate Löffler, Carola Hein, and Victor de Boer.

\bibliographystyle{IEEEtran}
\bibliography{main}
%



\end{document}